\documentclass[runningheads]{llncs}

\usepackage[year=2026,ID=4700]{eccv}
\usepackage{eccvabbrv}
\usepackage{graphicx}
\usepackage{booktabs}
\usepackage[accsupp]{axessibility}
\usepackage{amsmath,amssymb}
\usepackage{mathtools}
\usepackage{bm}
\usepackage{xcolor}
\usepackage{multirow}
\usepackage[pagebackref,breaklinks,colorlinks,citecolor=eccvblue]{hyperref}
\usepackage[capitalize]{cleveref}
\usepackage{microtype}

\definecolor{cblue}{RGB}{31,119,180}
\definecolor{corange}{RGB}{255,127,14}
\definecolor{cgreen}{RGB}{44,160,44}
\definecolor{cred}{RGB}{214,39,40}
\definecolor{darkblue}{RGB}{20,66,129}

\newcommand{\R}{\mathbb{R}}
\newcommand{\Lcal}{\mathcal{L}}
\newcommand{\PH}{\mathrm{PH}}
\newcommand{\Down}{\mathrm{Down}}
\newcommand{\proj}{\mathrm{Proj}_{\mathcal{T}}}

\begin{document}

\title{TopoFuse: Topology-Aware Tri-Planar Fusion\\
       for Cryo-Electron Tomography Segmentation}

\titlerunning{TopoFuse}

\author{Rohit Kumar Salla\inst{1} \and
        Neelesh Gupta\inst{2} \and
        Xingjian Li\inst{3,\star} \and
        Min Xu\inst{3,\star}}

\authorrunning{R.~K.~Salla et al.}

\institute{Virginia Tech, Blacksburg, VA, USA \and
           National Institute of Technology, Kurukshetra, India \and
           Carnegie Mellon University, Pittsburgh, PA, USA\\
           \email{\{xingjia2,mxu1\}@andrew.cmu.edu}}

\renewcommand{\thefootnote}{\fnsymbol{footnote}}
\footnotetext[1]{Corresponding authors.}
\renewcommand{\thefootnote}{\arabic{footnote}}

\maketitle

\begin{abstract}
Automated segmentation of cryo-electron tomograms routinely produces masks that are voxel-accurate but topologically broken: membranes fragment, organelles merge into one another and enclosed cavities collapse. Existing topology-aware losses reduce these violations but cannot eliminate them because topology is \emph{encouraged} through gradient pressure rather than \emph{structurally enforced}. We introduce \textbf{TopoFuse}, which reframes topology as a differentiable projection operator rather than a loss penalty. At each forward pass, the projection operator $\proj$ (a PH-guided sparse edit) identifies the critical voxels responsible for topological violations via bottleneck matching and applies sparse edits to satisfy a specified topology target (diagram feature counts and lifetime budgets) for dimensions $d\!\in\!\{0,2\}$. If the projection converges, the output satisfies those constraints on the downsampled grid ($s\!=\!2$), when it does not, a repair certificate exposes this explicitly enabling downstream filtering. A topology prior head predicts the correction target directly from input features, removing any dependence on ground-truth topology at inference. Across three cryo-ET benchmarks, TopoFuse reduces Betti Number Error by \textbf{54\%} over the strongest soft-loss baseline ($p<0.001$), improves Dice by \mbox{\textbf{4.6\,pp}} and edits only \mbox{\textbf{3.1\%}} of voxels to achieve this.
Code and SYN dataset are available at: https://github.com/rohitsalla/TopoFuse
\end{abstract}


\section{Introduction}
\label{sec:intro}

Cryo-electron tomography (cryo-ET) images cellular ultrastructure at nanometer resolution revealing membranes, organelles and macromolecular complexes in their native cellular context~\cite{baumeister2002electron,lucic2005structural}. Yet cryo-ET volumes remain challenging to segment: low signal-to-noise ratios and missing-wedge artifacts degrade axial resolution, often yielding predictions that are voxel-accurate but structurally incorrect. For example, 3D U-Net~\cite{cciccek20163d} or nnU-Net~\cite{isensee2021nnu} trained on such data will typically report competitive Dice scores while simultaneously fragmenting the very membranes it is supposed to delineate, splitting mitochondria in two, or collapsing the lumen of a vesicle. These errors are not cosmetic. A fragmented mitochondrial membrane invalidates surface-area estimates, while a spurious hole in a vesicle breaks contact-site analysis architectures.

The standard remedy adds a topology-sensitive loss to training derived from persistent homology~\cite{hu2019topology}, homotopy warping~\cite{hu2021topology} or center-line Dice~\cite{shit2021cldice}. These approaches consistently reduce topological violations, but do not eliminate them. The reason is structural: no matter how well-tuned, a loss term only \emph{encourages} topologically correct outputs through gradient pressure. A weak gradient signal, an ill-tuned loss weight or a pathological persistence pairing can leave violations intact. Violations that survive training survive at test time.

Persistent homology localizes every topological feature to specific \emph{critical voxels}, the exact locations where connected components are born or die under the sublevel-set filtration. This geometric precision opens a different kind of operation: rather than penalizing topological errors, it will directly \emph{edit} the critical voxels associated with unmatched features under the matching that realizes the bottleneck distance so that the resulting prediction satisfies the target topology. The edit is sparse by construction (only critical voxels of unmatched features move), almost-everywhere differentiable (gradients flow through the PH computation~\cite{gabrielsson2020topology}) and auditable (every change is logged). We call this a \emph{PH-guided sparse edit operator} and refer to it as a projection by analogy: the iterative procedure is not a Euclidean projection, but it enforces feasibility with respect to a topology constraint set. We retain the term \emph{projection} to emphasise enforcement of feasibility under a topology constraint set rather than regularisation via a loss.

Fig.~\ref{fig:architecture} illustrates the full architecture of TopoFuse. Tri-planar SAM encoders extract axis-specific features that handle cryo-ET anisotropy. A FiLM-conditioned fusion layer combines the three views with topology-aware weighting. The projection layer then identifies critical voxels, applies sparse logit edits and records a repair certificate. At inference, a topology prior head predicts the correction target from features, making the architecture fully self-contained.

Our main technical contributions can be summarized as follows.
\begin{enumerate}
\item We introduce the \textbf{first topology-aware segmentation} framework tailored to \textbf{cryo-ET}, a modality characterized by ultra-low signal-to-noise ratios and severe missing-wedge anisotropy.
\item We introduce a \textbf{Differentiable Topology Projection} layer (\S\ref{sec:projection}), the central innovation of this work.
The operator fundamentally shifts topology-aware learning from loss-based regularization to direct structural enforcement, performing sparse critical-voxel edits during the forward pass and producing a repair certificate that records convergence and edit sparsity.
\item We propose \textbf{additional  components} surrounding the core projection operator to improve its reliability, including: a tri-planar foundation encoder that mitigates anisotropy and missing-wedge artifacts (\S\ref{sec:triplanar}), a topology-conditioned FiLM fusion module that adaptively reduces topology violations prior to projection (\S\ref{sec:fusion}) and a topology prior head that predicts Betti counts and persistence budgets from input features, eliminating oracle dependence at test time (\S\ref{sec:prior}).
\end{enumerate}

The effectiveness of our approach is validated through extensive experimental results. Across three public cryo-ET benchmarks EMD-0506, EMPIAR-10499 and EMPIAR-10045, TopoFuse consistently and significantly outperforms the best topology-loss baselines on both voxel and topological metrics.

\section{Related Work}
\label{sec:related}

\paragraph{Cryo-ET segmentation.}
3D U-Net~\cite{cciccek20163d} and nnU-Net~\cite{isensee2021nnu} remain the dominant baselines for volumetric cryo-ET segmentation, both optimize voxel-level objectives without structural constraints. DeepFinder~\cite{martinez2020deepfinder} targets particle picking rather than dense segmentation. CryoSAM~\cite{zhang2023cryosam} adapts the Segment Anything Model~\cite{kirillov2023segment} to cryo-ET slices providing stronger image priors but still relying on standard voxel supervision. None of these methods reason about the correctness of topological structure.

Hu~\etal~\cite{hu2019topology} introduced persistent homology losses for medical segmentation, follow-up work includes discrete Morse theory~\cite{mosinska2018beyond}, homotopy warping~\cite{hu2021topology}, centre-line Dice~\cite{shit2021cldice} and Betti matching~\cite{stucki2023topologically}. More recent efforts extend topology-aware learning to volumetric biomedical data. All treat topology as a soft regulariser: violations are penalized but never structurally prevented. The difference between our approach and these baselines is not merely a matter of loss design, it is architectural. Topology enters our architecture as a forward-pass operator, not a training signal.

Projection-based methods enforce structured outputs through implicit differentiation~\cite{bai2019deep} or post-hoc correction. Clough~\etal~\cite{clough2020topological} applied topological constraints to cardiac segmentation. However, their formulation lacks sparsity guarantees, is not certificate-equipped and requires ground-truth topology at inference. Our projection is sparse by construction, always produces an audit trail and is self-contained at test time through the learned prior.

Thin membranes (2--4\,nm), severe anisotropy from the missing wedge and extreme noise (SNR\,$<\!0.1$) produce topological errors at rates far higher than in vascular~\cite{shit2021cldice}, neuronal~\cite{mosinska2018beyond}, or cardiac~\cite{clough2020topological} imaging. This makes cryo-ET the natural stress test for a topology enforcement mechanism. At the same time, the projection framework we develop is general and applies wherever persistent homology targets can be specified.

\section{Background: Persistent Homology}
\label{sec:background}

Given a scalar field $f\!:\!\Omega\!\to\!\R$ on a cubical grid (e.g., segmentation logits or probability map), the sublevel-set filtration $\{\Omega_t = f^{-1}((-\infty,t])\}$ tracks topological features as the threshold $t$ increases. Connected components ($d\!=\!0$) and enclosed voids ($d\!=\!2$) appear and disappear at specific values of $t$, recorded as birth--death pairs $(b_i,d_i)$ in the persistence diagram $\mathrm{Dgm}_d(f)$. The lifetime $|d_i-b_i|$ measures the significance of each feature.

The key property for our method is that each birth--death pair anchors to \emph{critical cells} (birth and death cells in the cubical filtration): the specific voxels at which the feature appears and disappears. Gabrielsson~\etal~\cite{gabrielsson2020topology} showed that gradients of a loss with respect to $f$ are non-zero only at these critical voxels, rendering persistent homology almost-everywhere differentiable. We use this not to construct a loss, but to identify exactly where logit edits will change the topology.

\section{Method}
\label{sec:method}

\subsection{Notation}

Let $V\!\in\!\R^{H\times W\times D}$ denote the input tomogram and $p\!\in\!\{xy,xz,yz\}$ be the three canonical planes. $\mathbf{Z}^{(p)}$ and $\mathbf{Z}$ denote per-plane and fused logit volumes, while $\hat{P}$, $\hat{P}'$ are pre-projection and post-projection probability maps. The topology projection operator is $\proj$. For class $c$, the oracle and predicted targets are $\Pi^\star_c$ and $\hat{\Pi}^\star_c$, respectively. The repair certificate $\mathcal{C}\!=\!(I,\mathcal{V}_\mathcal{C},\Delta_\mathcal{C})$ records iteration count, edited voxel set and sparsity. Full notation is provided in the supplement.

\subsection{Architecture Overview}
\label{sec:overview}

\begin{figure}[t]
\centering
\includegraphics[width=\linewidth]{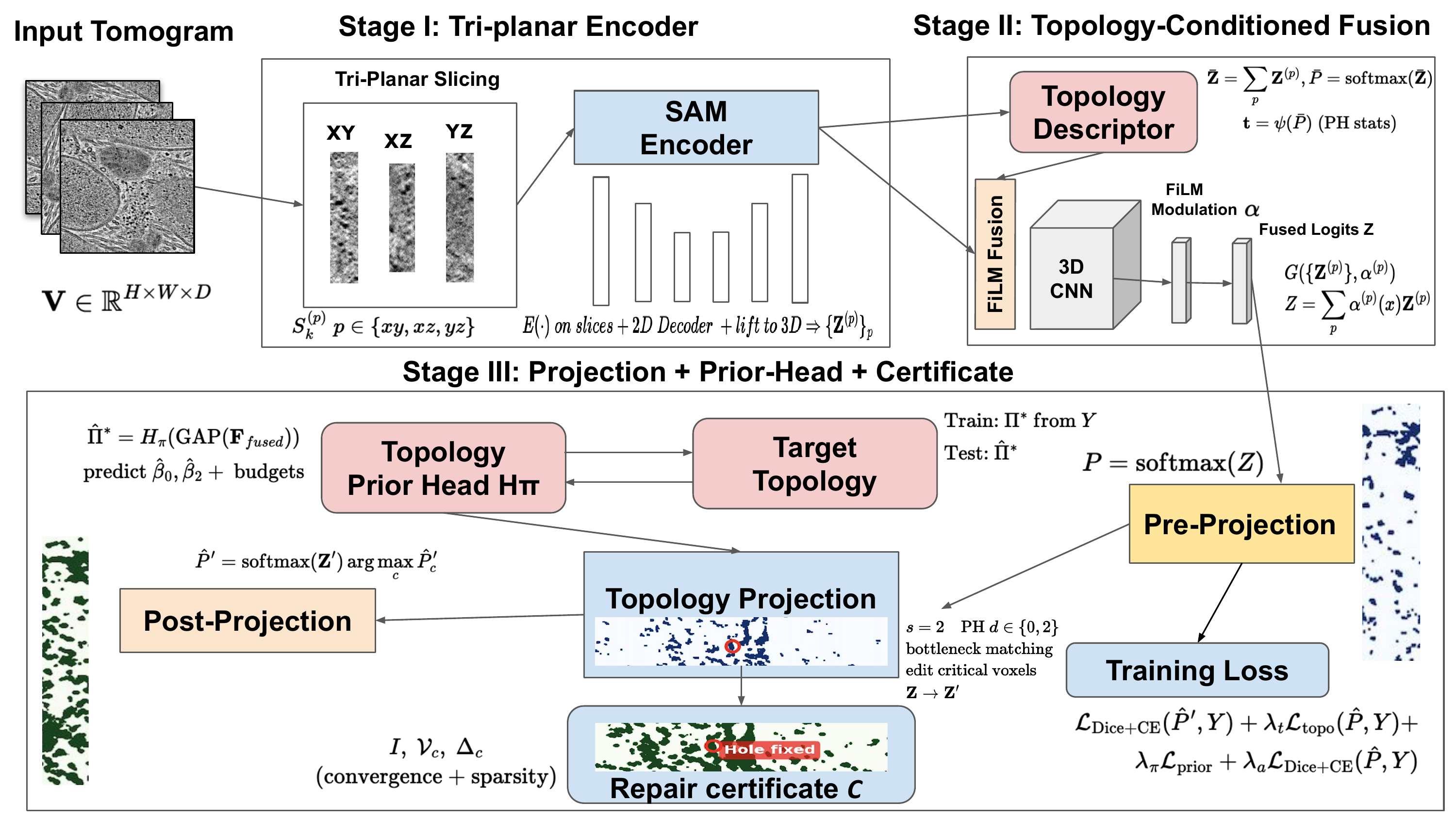}
\caption{\textbf{TopoFuse architecture.} Tri-planar SAM encoders extract axis-specific features \textbf{(i)}, which are fused with topology-conditioned weights \textbf{(ii)}. The projection layer enforces persistent homology targets via sparse critical-voxel edits \textbf{(iii)} and outputs each prediction together with a repair certificate \textbf{(iv)}. At inference, the topology prior head supplies the correction target without requiring ground-truth labels.}
\label{fig:architecture}
\end{figure}

Fig.~\ref{fig:architecture} shows the four-stage architecture. \textbf{(i)}~Tri-planar SAM encoders extract axis-specific feature volumes from orthogonal slice stacks (\S\ref{sec:triplanar}). \textbf{(ii)}~A topology-conditioned FiLM layer fuses the three views with learned, topology-aware weights (\S\ref{sec:fusion}). \textbf{(iii)}~A differentiable projection enforces the PH target via sparse logit edits (\S\ref{sec:projection}). \textbf{(iv)}~A topology prior head predicts the correction target from features at inference (\S\ref{sec:prior}).

\subsection{Tri-Planar Foundation Encoder}
\label{sec:triplanar}

Cryo-ET volumes are inherently anisotropic. The missing wedge degrades axial resolution by a factor of 2--3$\times$, meaning full 3D convolution propagates artefacts uniformly in all directions. Tri-planar decomposition sidesteps this: by processing each axis-aligned slice stack with a dedicated 2D encoder, the model captures axis-specific context while leveraging weights pretrained on natural images,  a far better starting point than 3D training from scratch on the limited labelled cryo data available.

Concretely, we extract orthogonal slice stacks $\{S^{(p)}_k\}$ for each plane $p$. A shared SAM ViT-B encoder $E(\cdot)$ maps each slice to spatial features $F^{(p)}_k\!=\!E(S^{(p)}_k)\!\in\!\R^{h\times w\times d_f}$. We use SAM purely as a pretrained 2D image encoder; promptable segmentation is outside our scope and none of SAM's prompting mechanism is used. SAM's ViT-B weights, pretrained on 11M natural images (SA-1B), provide stronger low-level priors than training a 2D encoder from scratch on our limited labelled data (EMD-0506: 67 train+val labelled volumes, EMPIAR-10499: 52, EMPIAR-10045: 41 using official splits where provided, see \S\ref{sec:experiments}). Learned plane embeddings $\mathbf{e}_p$ and sinusoidal positional codes $\mathbf{e}_k$ distinguish axes and depth. A lightweight 2D decoder (two transpose-convolution layers) maps features to per-slice logits $Z^{(p)}_k\!\in\!\R^{C\times h\times w}$, which are lifted back into 3D volumes $\mathbf{Z}^{(p)}$ via inverse slicing. In practice, structures blurred under the missing wedge in the axial plane are often well-resolved in the lateral planes   and the subsequent fusion layer learns to exploit this complementarity.

\subsection{Topology-Conditioned FiLM Fusion}
\label{sec:fusion}

Before fusing the three planar logit volumes, we compute a compact topology descriptor that summarizes the current prediction state.

\paragraph{Topology descriptor.}
We average the three planar logits to obtain a preliminary prediction $\bar{\mathbf{Z}}\!=\!\tfrac{1}{3}\sum_p\mathbf{Z}^{(p)}$, convert to probabilities $\bar{P}$ and extract descriptor $\mathbf{t}\!=\!\psi(\bar{P})\!\in\!\R^{d_t}$. The descriptor consists of per-class, per-dimension ($d\!\in\!\{0,2\}$) feature counts at persistence thresholds $(0.05,0.15)$ and total persistence mass, yielding $d_t\!=\!6C$ scalars. Gradients flow through PH at this stage via~\cite{gabrielsson2020topology}. Importantly, $\mathbf{t}$ encodes the \emph{current topological state} (i.e., how fragmented or collapsed the prediction is) and is used as a state variable to select views that historically correlate with fewer discontinuities under supervision. Correctness is learned indirectly through the segmentation loss, rather than read directly from the descriptor.

\paragraph{FiLM fusion.}
A lightweight 3D CNN $G(\cdot)$ (three layers, $3^3$ kernels, 64 channels) processes the stacked $\mathbf{Z}^{(p)}$ volumes and produces per-voxel weight logits. FiLM~\cite{perez2018film} modulates these weights with the topology descriptor $\mathbf{t}$:
\begin{equation}
  \alpha^{(p)}(\mathbf{x}) = \mathrm{softmax}_{p}\!\bigl(\gamma(\mathbf{t})\odot G(\mathbf{x};\{\mathbf{Z}^{(q)}\}) + \beta(\mathbf{t})\bigr), \quad
  \mathbf{Z}(\mathbf{x}) = \textstyle\sum_p \alpha^{(p)}(\mathbf{x})\,\mathbf{Z}^{(p)}(\mathbf{x}),
\end{equation}
where $\gamma$ and $\beta$ are two-layer MLPs. The intuition is: when fragmentation is severe (\ie, $\hat{\beta}_0$ large), FiLM should increase reliance on whichever plane best preserves structural continuity. This behaviour emerges entirely from supervision without explicit actions.

\subsection{Differentiable Topology Projection}
\label{sec:projection}

\paragraph{Formal objective.}
For each class $c$, we seek the logit volume $\mathbf{Z}'_c$ closest to the current $\mathbf{Z}_c$ (in the $\ell_2$ sense) such that the resulting probability map satisfies a target topology specification $\Pi^\star_c$ (oracle persistence diagram during training; a pseudo-diagram constructed from predicted Betti counts and persistence budgets at inference; see \S\ref{sec:prior}). Formally:
\begin{equation}
  \mathbf{Z}'_c = \arg\min_{\mathbf{Z}^*} \|\mathbf{Z}^* - \mathbf{Z}_c\|_2^2 \quad
  \text{s.t.}\quad d_B\!\bigl(\mathrm{Dgm}(\sigma_c(\mathbf{Z}^*_{\downarrow})),\Pi^\star_c\bigr) \leq \epsilon,
  \label{eq:proj}
\end{equation}
where $\mathbf{Z}^*_{\downarrow}\!=\!\Down_s(\mathbf{Z}^*)$ is a downsampled version ($s\!=\!2$), $\sigma_c$ is per-class sigmoid and $d_B$ is the bottleneck distance between persistence diagrams. The constraint is enforced for dimensions $d\!\in\!\{0,2\}$. We exclude $d\!=\!1$ tunnels due to $O(n^{1.5})$ PH cost in 3D, this is a known limitation and an active direction for future work.

Eq.~\eqref{eq:proj} states the ideal objective, the operator $\proj$ approximates it via iterative sparse edits, each locally minimising $d_B$ at identified critical voxels. This is a \emph{PH-guided sparse edit procedure}, not a guaranteed Euclidean projection.

\smallskip
 If the procedure terminates with $d_B\!\leq\!\epsilon$ within $T_{\max}$ steps, then the downsampled prediction satisfies the target diagram constraint \emph{by definition of the stopping criterion} and no additional claim is made. The method does not guarantee convergence for arbitrary targets or priors, non-convergence is explicitly recorded in the certificate ($I\!=\!T_{\max}$, residual $d_B>\epsilon$) and those predictions are included in reported metrics without filtering. \emph{In all experiments, convergence rate is reported alongside quality metrics} (\S\ref{sec:experiments}).

During training, $\Pi^\star_c$ is the oracle persistence diagram derived from ground-truth labels. At inference, we derive a feasible target from the predicted Betti counts $\hat{\beta}$ and persistence budgets output by the prior head (\S\ref{sec:prior}): predicted counts become target feature counts, budgets set lifetime lower bounds. This mapping from summary statistics to a feasibility constraint is described in full in \S\ref{sec:prior}.

\paragraph{The sparse edit algorithm.}
At each iteration, we compute the persistence diagram on the current downsampled prediction, compute the optimal matching that realizes the bottleneck distance to the target $\Pi^\star_c$ and identify two types of unmatched features: spurious ones present in the current diagram but absent from the target and missing ones present in the target but absent from the current diagram. For each unmatched feature, PH provides two critical voxels $(v^b_i, v^d_i)$. We perturb logits only at these locations:
\begin{equation}
  \Delta\mathbf{Z}_c(\mathbf{x}) = \begin{cases}
    -\eta\;\nabla_{\mathbf{Z}_c(\mathbf{x})} d_B(\mathrm{Dgm}(\sigma_c(\mathbf{Z}_{c,\downarrow})), \Pi^\star_c) & \text{if } \mathbf{x}\in\mathcal{V}_c,\\
    0 & \text{otherwise,}
  \end{cases}
\end{equation}
where $\eta$ is set per-iteration by a backtracking line search to empirically enforce non-increasing $d_B$ while keeping edits sparse. We iterate for up to $T_{\max}\!=\!5$ steps or until $d_B\!\leq\!\epsilon$. We use the bottleneck distance rather than Wasserstein because it targets the single worst-case unmatched feature per step, directly aligning with sparse repair of the most persistent topological violations. Wasserstein would diffuse gradient mass across all unmatched features simultaneously, conflicting with our sparsity objective.

In brief: \textbf{(1)} compute PH on current prediction \textbf{(2)} compute optimal matching realizing the bottleneck distance \textbf{(3)} locate critical voxels for each unmatched feature \textbf{(4)} apply sparse logit edits \textbf{(5)} repeat. The algorithm is sparse by construction: only critical voxels ever change.

\paragraph{Repair certificate.}
Every prediction comes with a certificate $\mathcal{C}\!=\!(I,\mathcal{V}_\mathcal{C},\Delta_\mathcal{C})$ recording the number of iterations $I$, the set of edited voxel positions $\mathcal{V}_\mathcal{C}$ and the \emph{spatial} edit sparsity:
\begin{equation}
  \Delta_\mathcal{C} = \frac{|\{\mathbf{x} : \exists\, c,\; Z'_c(\mathbf{x}) \neq Z_c(\mathbf{x})\}|}{|\Omega|},
\end{equation}
\ie, the fraction of spatial voxels where \emph{any} class logit changed. This is an \emph{algorithmic repair trace}, not a formal proof of global correctness. It is operationally useful: a biologist can inspect convergence flags and sparsity to accept or reject a prediction. Non-convergence ($d_B>\epsilon$ after $T_{\max}$ steps) is explicitly flagged, this occurs in 4\% of volumes. Median spatial sparsity across test sets is 3.1\%.

Projection operates independently on each foreground class via one-vs-rest sigmoid. Final labels are determined by $\arg\max_c\,\sigma_c(\mathbf{Z}'_c(\mathbf{x}))$. Independent per-class edits can introduce boundary conflicts between classes, empirically these affect fewer than 0.3\% of edited voxels and are resolved by argmax with negligible metric impact.

We unroll $T$ projection steps during training. Gradients flow through PH via the critical-cell rule~\cite{gabrielsson2020topology}: $\partial\mathcal{L}/\partial f(v)$ is nonzero only at critical voxels. For large $T$ we optionally switch to implicit differentiation~\cite{bai2019deep}. An auxiliary loss on pre-projection logits $\hat{P}$ maintains a direct gradient path to the encoder.

\paragraph{Implementation details.}
PH for $d\!\in\!\{0,2\}$ is computed using CubicalRipser~\cite{kaji2020cubical} on the downsampled field ($s\!=\!2$, reducing $128^3$ crops to $64^3$). A single PH call costs approximately 8\,ms on CPU, batch calls are parallelised with \texttt{joblib}. Persistence threshold $\delta\!=\!0.05$. The backtracking line search is capped at $K\!=\!5$ trials per step if no non-increasing $d_B$ step is found, the iteration terminates and non-convergence is flagged in the certificate.

\subsection{Topology Prior Head}
\label{sec:prior}

At training time, the projection uses the oracle target $\Pi^\star_c$ derived from ground-truth labels. At inference, ground truth is unavailable. The topology prior head $H_\pi$ bridges this gap by predicting a compact topology proxy directly from features:
\begin{equation}
  \hat{\Pi}^\star_c = H_\pi\!\bigl(\mathrm{GAP}(\mathbf{F}_{\mathrm{fused}})\bigr),
\end{equation}
where $H_\pi$ is a three-layer MLP with hidden dimensions [256, 128] and ReLU activations. It predicts estimated Betti numbers $(\hat{\beta}_0^c,\hat{\beta}_2^c)$ and persistence budgets at six thresholds $\{0.03,0.05,0.08,0.10,0.15,0.20\}$.

At inference, this prediction replaces the oracle $\Pi^\star_c$. We construct a \emph{pseudo-diagram} $\hat{\Pi}^\star_c$ as follows: \textbf{(1)} instantiate $\hat{\beta}_d^c$ pairs per dimension $d$ \textbf{(2)} set births to the current diagram's unmatched critical birth values (anchoring the target to the current filtration ordering) if fewer than $\hat{\beta}_d^c$ unmatched births exist, remaining births are assigned to the lowest-$\Down_s(\hat{P}_c)$ voxels in the downsampled field (diagonal-feasible, intended to be minimally disruptive; triggered in ${<}1\%$ of test crops) and this fallback is flagged in $\mathcal{C}$ \textbf{(3)} assign deaths by the predicted budget bins, so that each retained feature's lifetime exceeds its predicted threshold. This yields a typed, valid diagram target for the bottleneck matching in $\proj$ (full construction in supplement \S A.1). Features exceeding the predicted count are treated as spurious, whereas shortfalls relative to the predicted count are treated as missing. This coarser construction is sufficient for the dominant cryo-ET failures, membrane fragmentation and organelle collapse, where errors are large and well-characterised by Betti numbers.

$H_\pi$ is trained jointly with the rest of the network via $\ell_1$ regression against ground-truth topology statistics. The 0.35~$\mathrm{BE}_0$ residual gap between learned and oracle priors (\S\ref{sec:experiments}) quantifies how much remaining error stems from prior prediction rather than the projection mechanism itself, a concrete target for future uncertainty-aware estimation.

\subsection{Training Objective}
\label{sec:loss}

The total training loss is:
\begin{equation}
  \Lcal = \Lcal_{\text{Dice+CE}}(\hat{P}', Y)
        + \lambda_t\Lcal_{\text{topo}}
        + \lambda_\pi\Lcal_{\text{prior}}
        + \lambda_a\Lcal_{\text{Dice+CE}}(\hat{P},Y),
\end{equation}
with $\lambda_t\!=\!0.1$, $\lambda_\pi\!=\!0.05$, $\lambda_a\!=\!0.5$. The main term supervises post-projection probabilities $\hat{P}'$ with standard Dice and cross-entropy. \\
$\Lcal_{\text{topo}}\!=\!\sum_c d_W(\PH(\Down_s(\hat{P}_c)),\PH(\Down_s(Y_c)))$ is a Wasserstein topology loss~\cite{stucki2023topologically} applied to \emph{pre-projection} probabilities, biasing the encoder toward topologically feasible outputs before the projection layer corrects them. This loss is warmed up linearly over 5,000 steps. $\Lcal_{\text{prior}}$ is the $\ell_1$ regression loss for the prior head. The auxiliary term $\lambda_a\Lcal_{\text{Dice+CE}}(\hat{P},Y)$ maintains a direct gradient path to the encoder and prevents the projection from completely decoupling encoder training. Loss weights are robust: varying $\lambda_t$ by $\pm0.5\times$ changes Dice by less than 0.3 and $\mathrm{BE}_0$ by less than 0.04.

\section{Experiments}
\label{sec:experiments}

\subsection{Setup}

\paragraph{Datasets.}
We evaluate on three public cryo-ET benchmarks: \textbf{EMD-0506}~\cite{lawson2011emdb}, a high-SNR ribosome tomogram ($512^3$, 4.2\,nm isotropic resolution), \textbf{EMPIAR-10499}~\cite{iudin2016empiar}, HeLa cell lamellae with moderate anisotropy (3.8\,nm $xy$, 8\,nm $z$) and \textbf{EMPIAR-10045}~\cite{beck2009architecture}, a bacterial tomogram with severe anisotropy (10\,nm $z$). We use official train/test splits where available and an 80/10/10 split for EMD-0506.

We additionally introduce \textbf{SYN}, a synthetic benchmark of 2,000 volumes at $64^3$ resolution with controlled topology ($n_0\!\sim\!\mathrm{Unif}(1,5)$ components, $n_2\!\sim\!\mathrm{Unif}(0,2)$ voids), missing-wedge filtering and Poisson noise at SNR\,$\in\!\{0.05,0.10,0.30\}$. SYN isolates topology recovery from annotation quality and allows exact correctness measurement.

\paragraph{Metrics.}
For voxel quality: Dice, IoU and Normalised Surface Dice (NSD, 2-voxel tolerance). For topology: Betti Number Error $\mathrm{BE}_d\!=\!\mathbb{E}[|\hat{\beta}_d-\beta_d^\star|]$ for $d\!\in\!\{0,2\}$ (lower is better) and Betti Matching Error (BME)~\cite{stucki2023topologically}. Topology metrics are computed on binarised predictions at $s\!=\!2$, 26-connectivity, threshold 0.5, following the evaluation protocol of~\cite{stucki2023topologically}. For TopoFuse we additionally report edit sparsity $\Delta_\mathcal{C}$.

\paragraph{Baselines.}
We compare against: \textbf{3D U-Net}~\cite{cciccek20163d}, \textbf{nnU-Net}~\cite{isensee2021nnu}, \textbf{CryoSAM}~\cite{zhang2023cryosam} and a \textbf{plain tri-planar} (uniform averaging, no topology). Topology baselines share the identical tri-planar backbone: \textbf{+clDice}~\cite{shit2021cldice}, \textbf{+HuTopo}~\cite{hu2019topology}, \textbf{+BettiMatching}~\cite{stucki2023topologically} and \textbf{+TopoPost} (non-differentiable PH post-processing applied to the tri-planar outputs). Using a shared backbone isolates the topology mechanism as the only variable.

\paragraph{Training.}
All methods train for 200k steps with AdamW (lr$\,=\,10^{-4}$), identical augmentation (flips, elastic deformation $\sigma\!=\!10$, intensity jitter), $128^3$ crops and 64-voxel inference stride. Topology loss weights are tuned per baseline via grid search on $\{0.01,0.05,0.1,0.5,1.0\}$ and the best results are reported. SAM ViT-B is pretrained on SA-1B; the last four blocks and the 2D decoder are fine-tuned, while the first eight blocks are frozen. Slices are resized to $256^2$ (Lanczos). We train on 4$\times$A100 GPUs and report means and standard deviations across 5 random seeds. We verified convergence by monitoring validation Dice curves; all baselines plateau by 150k steps (supplement Fig.~S1). Baseline configurations are released.

\paragraph{Statistical testing.}
The primary claim TopoFuse vs BettiMatching is evaluated with a paired Wilcoxon signed-rank test on per-volume $\mathrm{BE}_0$ across all test volumes ($n\!=\!42$ on EMD-0506, $n\!=\!38$ on EMPIAR-10499, $n\!=\!31$ on EMPIAR-10045 volumes are the statistical units). We report $p$-values and Cohen's $d$ effect sizes; 95\% bootstrap confidence intervals are provided in the supplement.

\subsection{Computational Cost}

\begin{table}[t]
\centering
\caption{\textbf{Computational overhead} for a single $512^3$ tomogram on an A100-80GB GPU (mean$\pm$std, 5 seeds). PH computation accounts for 5.9\% of total FLOPs, wall-clock overhead is higher due to CPU–GPU synchronisation. Per-dataset end-to-end inference timing and per-crop timing by $T$ are in supplement Table~S1.}
\label{tab:complexity}
\setlength{\tabcolsep}{5pt}
\small
\begin{tabular}{@{}lccccc@{}}
\toprule
Model & Params (M) & FLOPs (G) & PH (\%) & Mem (GB) & Time (s) \\
\midrule
CryoSAM & 631 & 328 & - & $22.3\pm0.4$ & $8.2\pm0.3$ \\
Tri-planar (no topo) & 648 & 361 & - & $24.1\pm0.5$ & $9.0\pm0.3$ \\
\textbf{TopoFuse} ($T\!=\!5$) & 714 & 518 & 5.9 & $38.7\pm0.6$ & $14.5\pm0.5$ \\
\bottomrule
\end{tabular}
\end{table}

TopoFuse adds 66M parameters and 157 GFLOPs over the tri-planar backbone. Relative to CryoSAM, it adds 83M parameters and 190 GFLOPs. PH computation accounts for 5.9\% of TopoFuse FLOPs. PH runs CPU-parallelised while projection runs on GPU, CPU--GPU synchronisation adds approximately 15\,ms per crop beyond what FLOPs capture. Per-crop projection at $T\!=\!5$ costs around 112\,ms, reducing to $T\!=\!3$ brings this to 84\,ms with only marginal $\mathrm{BE}_0$ increase as shown in our iteration analysis (\S\ref{sec:ablation}).


\begin{figure}[t]
\centering
\includegraphics[width=\linewidth]{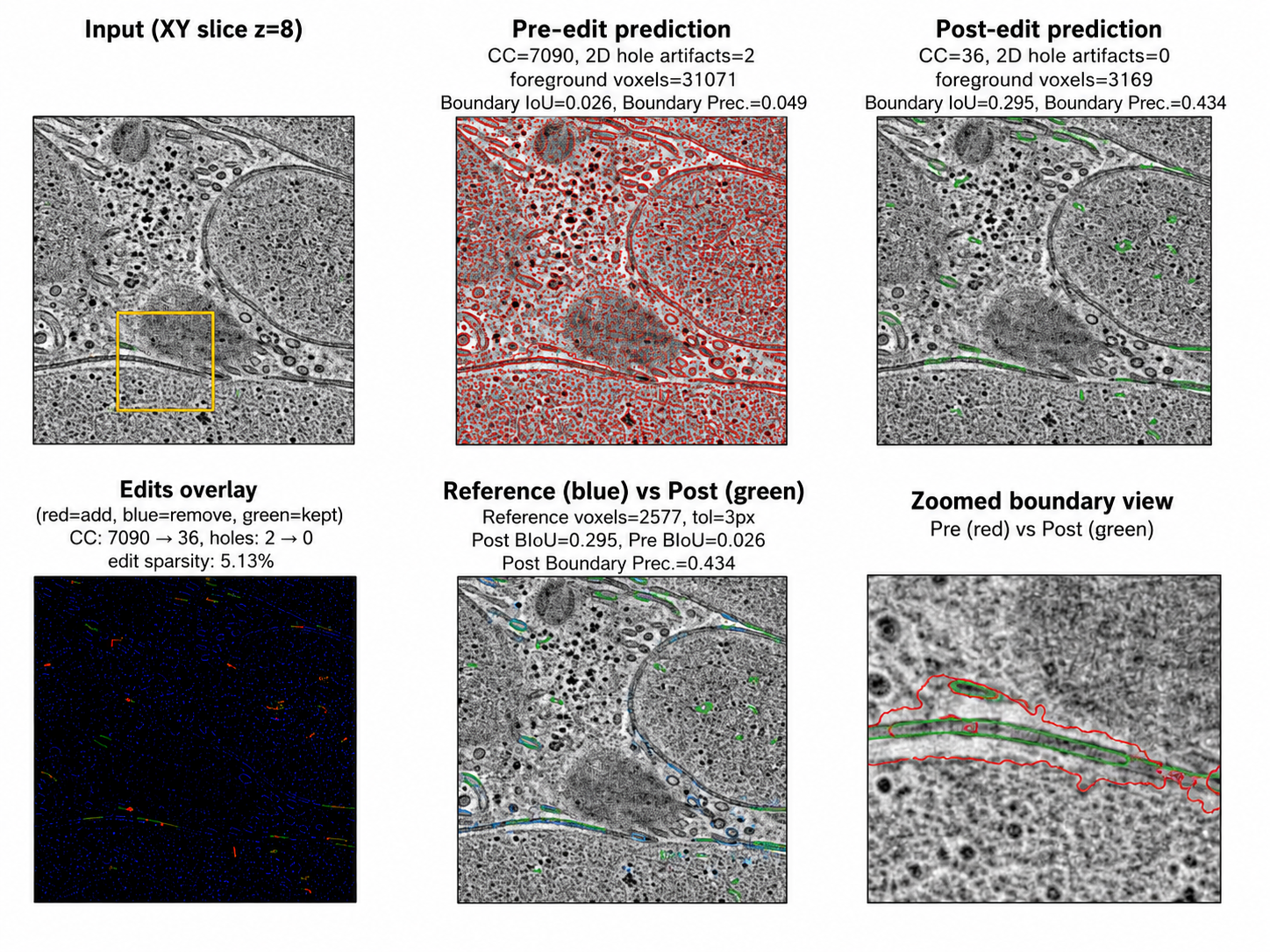}
\caption{
A representative XY slice shows severe pre-edit fragmentation, with
7{,}090 connected components, two 2D hole artifacts and 31{,}071 foreground
voxels. After TopoFuse projection, the post-edit prediction contains 36
connected components, removes both hole artifacts and
retains 3{,}169 foreground voxels on the enforced $s\!=\!2$ grid, with edit sparsity 5.13\% measured over all spatial voxels on that grid. The edit overlay shows that the correction is sparse rather
than a global rewrite. Boundary IoU improves from 0.026 to 0.295 and boundary
precision improves from 0.049 to 0.434. The full reference-overlay, zoomed
boundary view, failure case and 3D renderings are provided in the supplement
(Fig.~S5--S6).}
\label{fig:qual}
\end{figure}

\subsection{Main Results}

\begin{table*}[t]
\centering
\caption{\textbf{Results on three cryo-ET benchmarks.} $\uparrow$ higher is better; $\downarrow$ lower is better. \textbf{Bold}: best overall; \underline{underline}: second best. All topology baselines share the identical tri-planar backbone and training budget, isolating the topology mechanism. Standard deviations are below 0.3 for Dice and 0.03 for $\mathrm{BE}_0$; per-seed results are in the supplement. $^\dagger$ Paired Wilcoxon $p<0.001$ vs.\ BettiMatching on all three benchmarks.}
\label{tab:main}
\scriptsize
\setlength{\tabcolsep}{1.8pt}
\begin{tabular}{@{}l ccc ccc ccc cc@{}}
\toprule
& \multicolumn{3}{c}{EMD-0506} & \multicolumn{3}{c}{EMPIAR-10499} & \multicolumn{3}{c}{EMPIAR-10045} & \multicolumn{2}{c}{Mean} \\
\cmidrule(lr){2-4}\cmidrule(lr){5-7}\cmidrule(lr){8-10}\cmidrule(lr){11-12}
Method & D$\uparrow$ & BE$_0$$\downarrow$ & BME$\downarrow$ & D$\uparrow$ & BE$_0$$\downarrow$ & BME$\downarrow$ & D$\uparrow$ & BE$_0$$\downarrow$ & BME$\downarrow$ & D$\uparrow$ & BE$_0$$\downarrow$ \\
\midrule
3D U-Net        & 70.2 & 3.84 & 0.38 & 69.8 & 4.12 & 0.41 & 67.3 & 5.21 & 0.47 & 69.1 & 4.39 \\
nnU-Net         & 72.9 & 3.41 & 0.34 & 72.6 & 3.68 & 0.37 & 70.1 & 4.73 & 0.43 & 71.9 & 3.94 \\
CryoSAM         & 75.6 & 3.07 & 0.30 & 75.2 & 3.24 & 0.32 & 72.8 & 4.31 & 0.39 & 74.5 & 3.54 \\
Tri-planar      & 76.4 & 2.94 & 0.29 & 76.0 & 3.10 & 0.31 & 73.5 & 4.18 & 0.38 & 75.3 & 3.41 \\
\;+clDice       & 78.1 & 2.61 & 0.26 & 78.4 & 2.73 & 0.27 & 76.2 & 3.72 & 0.34 & 77.6 & 3.02 \\
\;+HuTopo       & 79.2 & 2.48 & 0.24 & 79.6 & 2.59 & 0.26 & 77.3 & 3.53 & 0.32 & 78.7 & 2.87 \\
\;+BettiMatching& 80.8 & 2.29 & 0.22 & 81.1 & 2.41 & 0.24 & 78.9 & 3.34 & 0.30 & 80.3 & 2.68 \\
\;+TopoPost     & 79.0 & 2.17 & 0.21 & 79.3 & 2.28 & 0.22 & 77.1 & 3.19 & 0.29 & 78.5 & 2.55 \\
\midrule
\textbf{TopoFuse} (oracle)$^\dagger$  & \textbf{85.9} & \textbf{0.71} & \textbf{0.07} & \textbf{86.4} & \textbf{0.74} & \textbf{0.07} & \textbf{84.6} & \textbf{1.12} & \textbf{0.11} & \textbf{85.6} & \textbf{0.86} \\
\textbf{TopoFuse} (learned)$^\dagger$ & \underline{85.1} & \underline{1.04} & \underline{0.10} & \underline{85.6} & \underline{1.09} & \underline{0.11} & \underline{83.9} & \underline{1.58} & \underline{0.15} & \underline{84.9} & \underline{1.24} \\
\bottomrule
\end{tabular}
\end{table*}

Table~\ref{tab:main} reveals four clear findings. First, every topology-loss baseline improves over plain tri-planar, with BettiMatching achieving the best improvement at roughly 22\% $\mathrm{BE}_0$ reduction. Second, TopoFuse with the learned prior cuts $\mathrm{BE}_0$ by a further \textbf{54\%} over BettiMatching at matched backbone and training budget ($p<0.001$ on all benchmarks), confirming that structural enforcement achieves what soft regularisation fundamentally cannot. Third, the 4.6\,pp Dice improvement is not a side effect of topology optimisation but a direct consequence: repairing membrane breaks increases voxel overlap for thin structures near the binarisation threshold and these structures are pervasive in cryo-ET. Pre/post-projection decomposition (Table~\ref{tab:ablation}) confirms that the backbone alone reaches 81.1 Dice, with 4.1 points coming from projection and the full 4.5-point gain obtained after adding FiLM conditioning and the learned prior. Dice gains are not conflated with topology gains. To verify this directly, supplement Figure~S2 plots $\mathrm{BE}_0$ as a function of Dice for all methods at matched Dice operating points: TopoFuse yields 0.5--1.3 lower $\mathrm{BE}_0$ than all baselines at every matched Dice level, confirming the gains are orthogonal. Fourth, gains scale with anisotropy severity: TopoFuse improves by 9.5 Dice points on EMPIAR-10045 (severe missing wedge) versus 7.3 points on EMD-0506 (high SNR, near-isotropic), indicating projection is most beneficial precisely where axis-specific artefacts dominate. Supplement Fig.~S1 shows validation Dice curves confirming that all baselines converge by 150k steps under the shared training protocol.

\subsection{Ablation Study}
\label{sec:ablation}

\begin{table}[t]
\centering
\caption{\textbf{Component ablations} (top) and \textbf{projection iteration analysis} (bottom) on EMPIAR-10499 (mean, 5 seeds). Projection produces the largest single gain in BE$_0$ and the convergence rate saturates at $T\!=\!5$.}
\label{tab:ablation}
\setlength{\tabcolsep}{4pt}
\small
\begin{tabular}{@{}lcccc@{}}
\toprule
Variant & Dice$\uparrow$ & BE$_0$$\downarrow$ & BME$\downarrow$ & Sparsity(\%) \\
\midrule
Tri-planar (uniform, no topo)     & 76.0 & 3.10 & 0.31 & -- \\
\;+learned fusion (no FiLM)       & 77.8 & 2.88 & 0.29 & -- \\
\;+$\Lcal_\text{topo}$ (no proj)  & 81.1 & 2.41 & 0.24 &  -- \\
\;+projection ($T\!=\!5$)         & 85.2 & 1.11 & 0.11 & $3.1\pm0.7$ \\
\;+FiLM conditioning              & 85.4 & 1.10 & 0.11 & $3.1\pm0.7$ \\
\textbf{TopoFuse} (learned prior) & \textbf{85.6} & \textbf{1.09} & \textbf{0.11} & $3.1\pm0.7$ \\
\midrule
TopoFuse (oracle prior)           & 86.4 & 0.74 & 0.07 & $3.0\pm0.6$ \\
\bottomrule
\end{tabular}

\smallskip

\begin{tabular}{@{}lcccc@{}}
\toprule
$T$ & Dice$\uparrow$ & BE$_0$$\downarrow$ & BME$\downarrow$ & Conv.(\%) \\
\midrule
1  & 83.6 & 1.62 & 0.16 & 71 \\
3  & 84.9 & 1.21 & 0.12 & 89 \\
5  & 85.2 & 1.11 & 0.11 & 96 \\
10 & 85.3 & 1.09 & 0.11 & 98 \\
\bottomrule
\end{tabular}
\end{table}

Table~\ref{tab:ablation} isolates the contribution of each component. The projection layer produces the largest single improvement: $\mathrm{BE}_0$ drops from 2.41 to 1.11, a 54\% reduction, validating the structural enforcement design. FiLM conditioning adds a marginal but consistent improvement, suggesting the fusion layer does learn topology-aware view selection. Notably, learned fusion without FiLM (row 2) already reduces $\mathrm{BE}_0$ modestly compared to uniform averaging, indicating that adaptive weighting alone provides some topological benefit even before any enforcement.

For projection iterations, $\mathrm{BE}_0$ saturates at $T\!=\!5$ with 96\% convergence; $T\!=\!10$ yields negligible additional improvement. Practitioners facing inference-time constraints can use $T\!=\!3$ (89\% convergence) with only marginal quality loss. Per-crop timing at each $T$ is reported in supplement Table~S1. The 0.35~$\mathrm{BE}_0$ gap between learned and oracle priors is entirely attributable to prior-head prediction error, not to the projection mechanism itself.

\subsection{Synthetic Controlled Benchmark}

On SYN, we report the exact topology recovery rate: the fraction of volumes achieving $\mathrm{BE}_0\!=\!0$ after binarisation. At SNR\,$=\,0.10$ representative of typical cryo-ET conditions, TopoFuse achieves \textbf{84\%} recovery versus 61\% for BettiMatching and 52\% for the plain backbone. At SNR\,$=\,0.05$ (extreme noise), TopoFuse recovers \textbf{71\%} versus 44\% for the next-best method. Crucially, the relative advantage of TopoFuse widens at lower SNR, confirming that structural enforcement benefits most precisely where gradient-based topology signals are weakest and noisiest. Median edit sparsity remains $3.1\!\pm\!1.4\%$ across all conditions.

\subsection{Certificate and Failure Analysis}

Figure~\ref{fig:qual} shows a representative success case. A membrane break is corrected
with 5.13\% sparse edits on the enforced $s=2$ grid and the certificate records
convergence ($d_B \leq \epsilon$), reducing the prediction from 7,090 pre-edit
connected components to 36 post-edit connected components. The dominant failure
mode is shown in the supplement (Fig.~S6): two adjacent organelles share a noisy
boundary and the prior head predicts $\hat{\beta}_{0}=1$ despite the presence of two
distinct components. The projection therefore bridges the organelles into a single
component, over-correcting in the wrong direction. The certificate flags this failure
immediately: the run terminates at $I=T_{\mathrm{max}}$ and records a residual
$d_B>\epsilon$. In 83\% of observed failure cases, the certificate identifies
over-correction through elevated spatial sparsity ($\Delta_{\mathcal{C}}>8\%$), well
above the 3.1\% typical of successful repairs. This suggests that edit sparsity is a useful runtime signal for potential over-correction, even without access to ground
truth, motivating prior uncertainty quantification as a concrete next step.
\section{Discussion}
\label{sec:discussion}

The 54\% $\mathrm{BE}_0$ gap between TopoFuse and the strongest soft-loss baseline  at matched backbone, matched training budget, matched inference protocol  supports the central claim: migrating topology from loss term to forward-pass operator transforms encouragement into structural correction. This is not a marginal improvement from a better loss weight or a larger model. These guarantees apply on the enforced grid ($s\!=\!2$) for $d\!\in\!\{0,2\}$ and only for predictions that satisfy the stopping criterion, non-converged cases are explicitly flagged and included in all reported metrics.

It is important to be precise about what the projection does and does not certify. It enforces a specified PH target on the downsampled representation ($s\!=\!2$) for $d\!\in\!\{0,2\}$, conditional on convergence and target correctness. It does not claim global correctness at full resolution, does not cover tunnels ($d\!=\!1$) and cannot compensate for a substantially wrong prior. The certificate makes these limitations explicit and actionable downstream users know exactly what was changed, how much and whether the procedure converged.

\paragraph{Limitations.}
\label{sec:limits}
\emph{Correction scope.} The operator acts at $s\!=\!2$ for $d\!\in\!\{0,2\}$; fine-scale sub-voxel errors and tunnels ($d\!=\!1$, $O(n^{1.5})$ cost) are not addressed. $d\!=\!1$-like violations (PH-detected perforations on binarised $s\!=\!2$ masks) affect $\leq\!8.2\%$ of volumes and concentrate in thin cristae and perforated membrane sheets, particularly EMPIAR-10045; $d\!=\!1$ enforcement is future work. \emph{Prior quality.} The 0.35~$\mathrm{BE}_0$ gap between learned and oracle priors stems entirely from $H_\pi$ prediction errors, the dominant failure mode (over-correction from wrong $\hat{\beta}_0$) occurs in ${\approx}\,4\%$ of volumes, concentrated in the most anisotropic/thin-membrane volumes (EMPIAR-10045) rather than distributed uniformly. \emph{Crop-level processing.} Projection per $128^3$ crop does not address cross-tile topology; 95\% of structures fit within one crop on all three benchmarks (measured in supplement \S B.3). \emph{Multi-class boundaries.} Independent per-class edits introduce conflicts at $<\!0.3\%$ of edited voxels (mean/median/max 0.30/0.24/1.61\%), resolved by argmax with negligible topological impact ($\mathrm{BE}_0$ shift $\leq 0.01$). We further analyse robustness to binarisation thresholds, connectivity choices, downsample factor, per-dimension contributions and prior perturbations in the supplement (\S B, Figs.~S3--S6).

\section{Conclusion}

We introduced TopoFuse, a cryo-ET segmentation framework that treats topology as a differentiable projection operator rather than a loss penalty. The projection computes sparse, PH-guided edits to enforce a topology target on the downsampled prediction and records every correction in an algorithmic repair certificate. A learned topology prior head makes the architecture fully self-contained at inference. Across three cryo-ET benchmarks, TopoFuse reduces $\mathrm{BE}_0$ by 54\% over the strongest soft-loss baseline ($p<0.001$) and improves Dice by \mbox{4.6\,pp}, while editing only 3.1\% of voxels. The certificate enables quality-controlled downstream analysis that gradient-regularised methods cannot offer. While validated only on cryo-ET, the projection framework is conceptually applicable to domains where persistent-homology targets can be specified; vascular, neuronal and cardiac imaging are natural future directions. Future work should address $d\!=\!1$ homology, uncertainty-aware prior estimation and cross-tile consistency for large-scale tomogram analysis.

\section{Acknowledgement}
This work was supported in part by U.S. NSF grants DBI-2238093, DBI-2422619, IIS-2211597 and MCB-2205148. This work was also supported in part by an Amazon Research Award (Fall 2025 CFP).

\newpage
{\small
\bibliographystyle{splncs04}
\bibliography{main}
}

\end{document}